\documentclass[10pt,twocolumn,letterpaper]{article}

\newif\ifarxiv
\arxivtrue

\ifarxiv
\usepackage[pagenumbers]{cvpr} %
\else
\usepackage{cvpr}              %
\fi

\def\MYTITLE{E-VLC: A Real-World Dataset for \\
Event-based Visible Light Communication And Localization}

\definecolor{cvprblue}{rgb}{0.21,0.49,0.74}
\ifarxiv
\usepackage[breaklinks,colorlinks,allcolors=cvprblue,bookmarks=true,bookmarksnumbered=true]{hyperref}
\hypersetup{
  pdftitle={\MYTITLE},
  pdfsubject={Computer Vision, Robotics},
  pdfauthor={Shintaro Shiba, Quan Kong, Norimasa Kobori},
  pdfkeywords={Event Cameras, Visible Light Communication, Optical Communication, Motion Estimation}
}
\usepackage[absolute]{textpos}
\else
\usepackage[pagebackref,breaklinks,colorlinks,allcolors=cvprblue]{hyperref}
\fi

\usepackage{amsmath,amssymb,amsfonts}
\usepackage{graphicx}
\usepackage{caption}
\usepackage{subcaption}
\usepackage{makecell}
\usepackage{booktabs}
\usepackage{siunitx}
\usepackage{adjustbox} %
\usepackage{array}
\usepackage{multirow}
\usepackage{rotating}
\usepackage{etoolbox}
\robustify\bfseries

\usepackage{url}
\usepackage{color}

\def\Lum{L}
\def\tref{t_\text{ref}} %
\def\pol{p} %

\def\cE{\mathcal{E}} %
\def\numEvents{N_e} %
\def\Warp{\mathbf{W}}

\def\bx{\mathbf{x}}
\def\bparams{\btheta}
\def\Rot{\mathtt{R}}

\def\pol{p}

\def\angvel{\boldsymbol{\omega}} %

\def\cN{\mathcal{N}} %
\def\bmu{\boldsymbol{\mu}} %

\def\IWE{I} %

\def\mId{\mathtt{Id}} %

\newcommand{\bnum}[1]{\bfseries #1}

\newcommand{\novalue}{{\textendash}}

\definecolor{light-gray}{gray}{0.6}

\usepackage[capitalize]{cleveref}
\crefname{section}{Sec.}{Secs.}
\Crefname{section}{Section}{Sections}
\Crefname{table}{Table}{Tables}
\crefname{table}{Tab.}{Tabs.}

\def\Lum{L}
\def\tref{t_\text{ref}} %
\def\pol{p} %

\def\cE{\mathcal{E}} %
\def\numEvents{N_e} %
\def\Warp{\mathbf{W}}

\def\bx{\mathbf{x}}
\def\bparams{\boldsymbol{\theta}}
\def\Rot{\mathtt{R}}

\def\pol{p}

\def\angvel{\boldsymbol{\omega}} %
\def\bphi{\boldsymbol{\phi}}
\def\bmu{\boldsymbol{\mu}}
\def\mId{\mathtt{Id}}
\def\IWE{I}

\def\cN{\mathcal{N}} %
\def\cT{\mathcal{T}} %

\def\baseF{f_{\text{b}}}

\usepackage[super]{nth}
\usepackage[sectionbib]{chapterbib}

\def\paperID{6} %
\def\confName{CVPR}
\def\confYear{2025}

\title{\MYTITLE}

\author{Shintaro Shiba \quad
Quan Kong \quad 
Norimasa Kobori \\
Woven by Toyota \\
}

\begin{document}
\ifarxiv
\definecolor{somegray}{gray}{0.5}
\newcommand{\darkgrayed}[1]{\textcolor{somegray}{#1}}
\begin{textblock}{11.5}(2.25, 0.8)  %
\begin{center}
\darkgrayed{This paper has been accepted for publication at the\\
IEEE Conference on Computer Vision and Pattern Recognition (CVPR) Workshops, Nashville, 2025.
\copyright IEEE}
\end{center}
\end{textblock}
\fi

\maketitle

\begin{abstract}

Optical communication using modulated LEDs (e.g., visible light communication) is an emerging application for event cameras, thanks to their high spatio-temporal resolutions.
Event cameras can be used simply to decode the LED signals and also to localize the camera relative to the LED marker positions.
However, there is no public dataset to benchmark the decoding and localization in various real-world settings.
We present, to the best of our knowledge, the first public dataset that consists of an event camera, a frame camera, and ground-truth poses that are precisely synchronized with hardware triggers.
It provides various camera motions with various sensitivities in different scene brightness settings, both indoor and outdoor.
Furthermore, we propose a novel method of localization that leverages the Contrast Maximization framework for motion estimation and compensation.
The detailed analysis and experimental results demonstrate the advantages of LED-based localization with events over the conventional AR-marker--based one with frames, as well as the efficacy of the proposed method in localization.
We hope that the proposed dataset serves as a future benchmark for both motion-related classical computer vision tasks and LED marker decoding tasks simultaneously,
paving the way to broadening applications of event cameras on mobile devices.
\textcolor{magenta}{\url{https://woven-visionai.github.io/evlc-dataset}}

\end{abstract}

\section{Introduction}
\label{sec:intro}

Optical communication from infrastructure to cameras, such as QR code and AR markers,
has a wide range of applications from mobile payment to localization in our daily lives.
These QR and AR markers are recognized mainly using conventional frame-based cameras,
however, recently active markers that modulate visible light (e.g., LED) at high frequencies
have gained attention thanks to event cameras that offer high spatio-temporal resolution.
However, due to the novelty of the camera,
there have been no public datasets or detailed analyses for such visible light communication (VLC).
Furthermore, it poses unique challenges for typical event-based vision methods to simultaneously process the high-frequency LED and motion data,
since the LEDs break the commonly-used brightness constancy assumption.
Prior work have shown that event cameras in VLC can be used both for receiving information and localization of moving markers (i.e., with a static camera) or a moving camera (i.e., with static markers).
However, the evaluation is challenging, especially where the camera moves, since it requires the ground truth (GT) poses.
\begin{figure}[t]
  \centering
  {{\includegraphics[width=\linewidth]{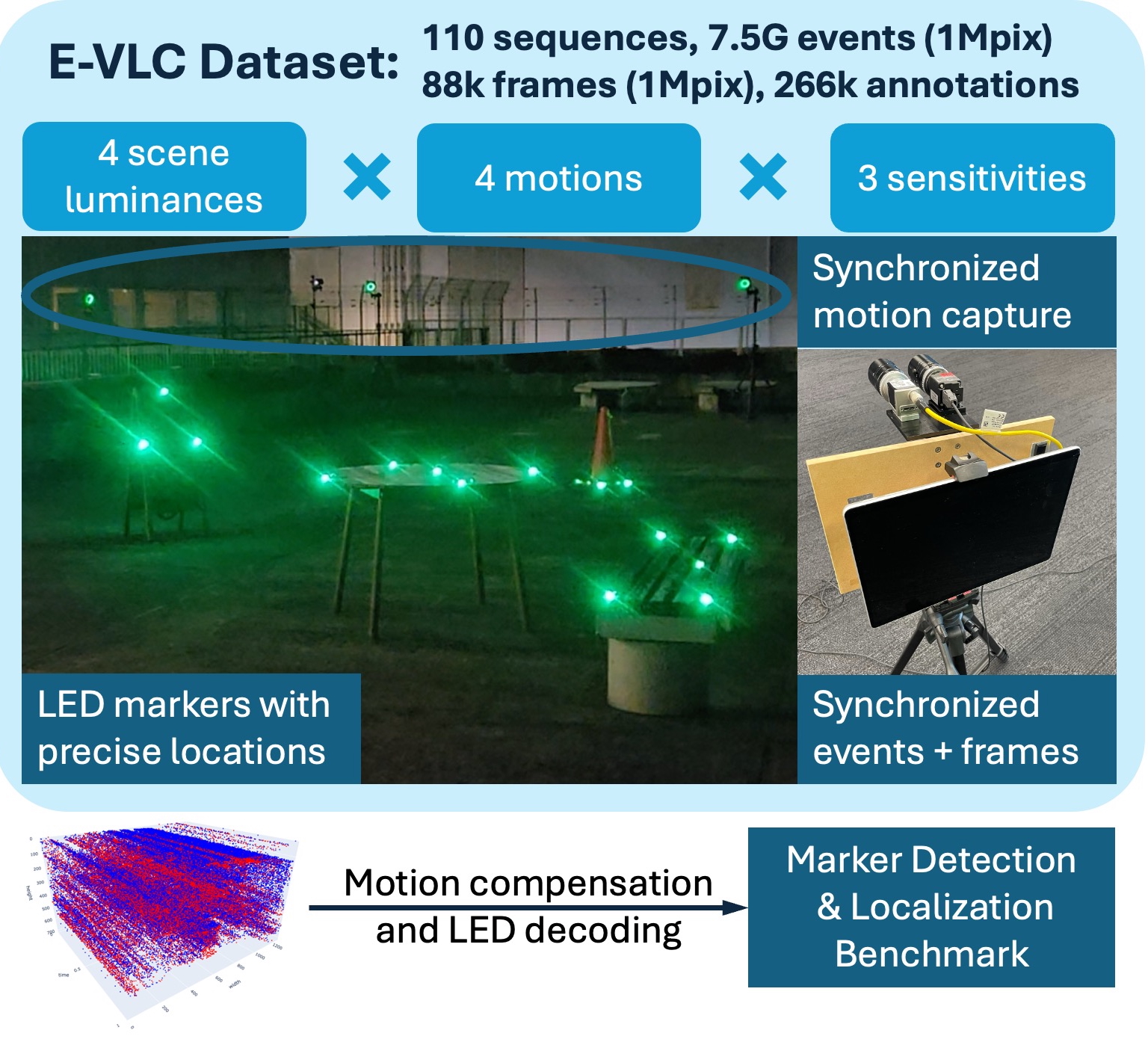}}}
\caption{We propose the E-VLC dataset and a novel method that compensates motion to improve the localization accuracy. The dataset consists of synchronized events, frames, and GT poses as well as bounding boxes, in various recording settings.}
\label{fig:eyeCatcher}
\vspace{-2ex}
\end{figure}

\begin{figure*}[t]
  \centering
  {{\includegraphics[clip,trim={0 7cm 0.4cm 0},width=\linewidth]{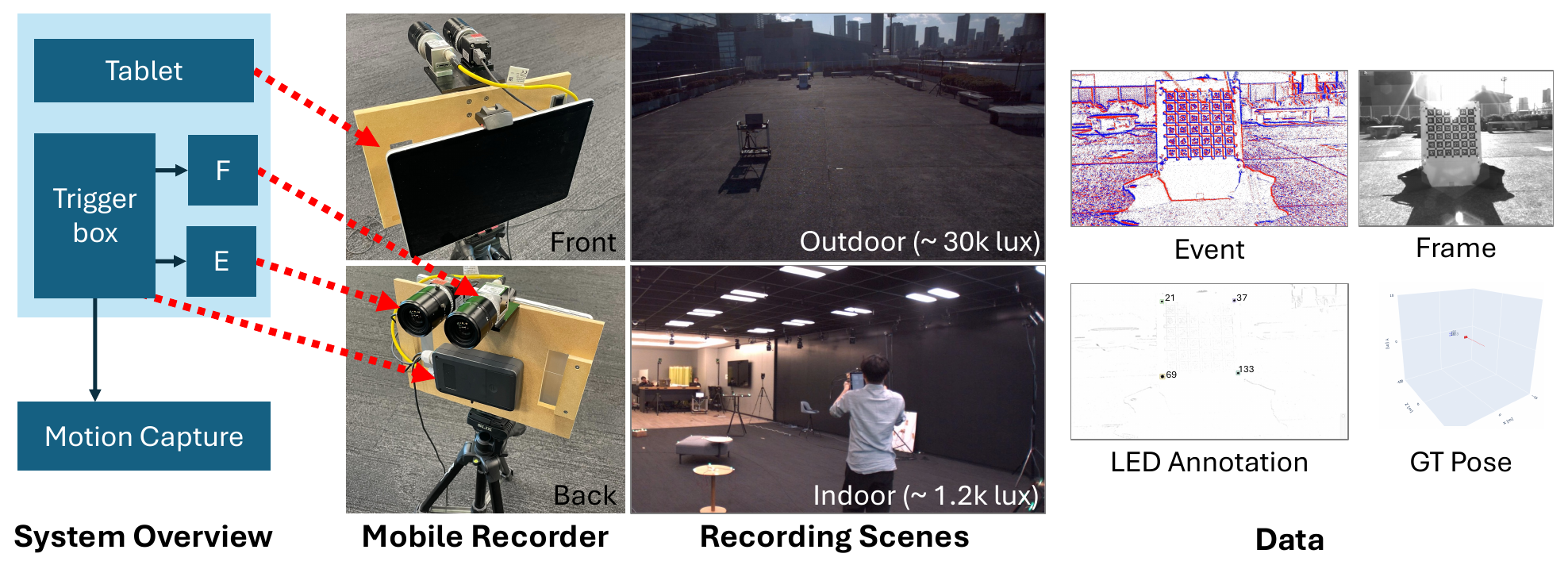}}}
\caption{Data recording setup.
The mobile recorder consists of two cameras ($1280\times1024$-pix frame and $1280\times720$-pix event) that are synchronized via a trigger box and a tablet. The trigger sends the signal to the external motion capture. We record the data both in outdoor and indoor environments for different scene luminance. Best viewed online.}
\label{fig:methodRecording}
\vspace{-2ex}
\end{figure*}

In this work, we present \emph{E-VLC}, the first event camera dataset that consists of
events and frames at high resolution ($1$-Mpixel) with ground-truth poses from motion capture, all of which are calibrated and accurately synchronized (\cref{fig:eyeCatcher}).
Precise synchronization is achieved by using an external trigger box that sends hardware triggers to the motion capture and the cameras on the mobile (hand-held) recording system.
The dataset offers various recording scenarios, including
($i$) different types of motion (static, rotation, translation, and dynamic),
($ii$) various scene luminance in both indoor and outdoor settings,
and ($iii$) different event camera sensitivities.
We also publish annotations of LED bounding boxes.

Detecting and decoding modulated LED signals from a moving event camera is challenging,
since such scenes include event data from moving edges and flickering LEDs.
Furthermore, due to the pixel displacements of the detected LEDs, the localization accuracy may deteriorate.
To this end, we propose a novel method that combines motion estimation and compensation for localization.
The experimental results show that
($i$) despite the lack of brightness constancy, the Contrast Maximization framework helps estimate motion in such scenarios
and ($ii$) the proposed motion compensation improves localization.
\begin{table}[t]
\centering
\caption{Comparison of different sensors for VLC.}
\setlength{\tabcolsep}{4pt}
\begin{tabular}{l*{2}{S}}
\toprule
& \text{Temporal resolution} & \text{Spatial resolution} 
\\ 
\midrule
\text{Photodiode} & \text{High (M--GHz)} & \text{Low} \\
\text{Frame camera} & \text{Low ($<100$Hz)} & \text{High} \\
\text{Event camera} & \textbf{High} \text{ ($<1$MHz)} & \textbf{High} \\
\bottomrule
\end{tabular}
\label{tab:vlccomparison}
\vspace{-2ex}
\end{table}

In summary, this work presents several distinctive contributions in event-based optical communication:
\begin{enumerate}
\item The first public dataset for VLC applications using event cameras. The dataset consists of precisely synchronized events, frames, ground-truth pose of LED markers and cameras, and bounding-box annotations for LEDs. It also includes various recording conditions (motion type, scene luminance, event camera sensitivity) to cover real-world applications.
\item Detailed analysis of the data under different settings, such as effective distance and comparison with frame-based methods. It helps the holistic understanding of the dataset that includes flickering LEDs and scene motion. %
\item A novel LED-based localization method that leverages motion compensation with Contrast Maximization. %
The experimental results show the proposed method successfully compensates motion and improves LED detection and localization. %
\end{enumerate}
With these contributions, this work will serve as a future benchmark for LED detection and LED-based localization tasks using event cameras.
We hope that the dataset under various practical scenarios will accelerate event camera applications, especially on edge devices such as AR glasses and mobiles, broadening the community \cite{Shiba25cvprw2}.

\sisetup{round-mode=places,round-precision=1}
\begin{table*}[t!]
\centering
\caption{Dataset summary.}
\setlength{\tabcolsep}{4pt}
\begin{tabular}{ll*{6}{S}}
\toprule
\text{Motion} & \text{Luminance [lx]} & \text{Sensitivity} & \text{Sequences} & \text{Duration [s]} & \text{Events} & \text{Frames} & \text{Annotations}
\\ 
\midrule
\text{Static} & \text{600, 1200, 30k} & \text{Low}  & \text{27} & 128.543 & \text{22M} & \text{2976} & \text{8928} \\
 & &  \text{Medium} & \text{23} & 71.686 & \text{82M} & \text{960} & \text{2880} \\
 & & \text{High} & \text{23} & 107.84599999999999 & \text{225M} & \text{2148} & \text{6444} \\
 \midrule
\text{Translation} & \text{600, 1200} & \text{Low}  & \text{6} & 378.37 & \text{41M} & \text{13629} & \text{40861} \\
& & \text{Medium} & \text{4} & 242.7 & \text{1129M} & \text{8005} & \text{23997} \\
& & \text{High} & \text{2} & 90.2 & \text{767M} & \text{3097} & \text{9284} \\
 \midrule
\text{Rotation}  & \text{600, 1200} & \text{Low} & \text{4} & 168.9 & \text{25M} & \text{6376} & \text{19128} \\
& & \text{Medium} & \text{4} & 208.7 & \text{1135M} & \text{6991} & \text{20973} \\
& & \text{High} &  \text{4} & 252 & \text{2031M} & \text{9092} & \text{27276} \\
 \midrule
\text{Dynamic}& \text{10, 600, 1200} & \text{Low} & \text{5} & 356.1 & \text{62M} & \text{13400} & \text{40171} \\
& & \text{Medium} & \text{5} & 386 & \text{983M} & \text{13954} & \text{41841} \\
& & \text{High} & \text{3} & 221 & \text{978M} & \text{8139} & \text{24404} \\
\midrule
Total  & \text{10, 600, 1200, 30k} & \text{L, M, H} & \text{110} & 2612 & \text{7.5G} & \text{88.7k} & \text{266.2k} \\ 
\bottomrule
\end{tabular}
\label{tab:dataset}
\end{table*}

\section{Related Work}
\label{sec:related}

Visible light communication (VLC) that utilizes modulated LEDs at high frequencies
has been known since the early 2000s \cite{Wook06ieice}.
Compared with wireless communication using radio frequency (RF),
VLC offers advantages such as high reliability and secure connections. %
However, the lack of suitable receivers prevents VLC from wide adaptation (\cref{tab:vlccomparison}).
Conventional frame-based cameras (CMOS/CCD) are not suitable for large data bandwidth,
because their temporal resolution (typically $10$--$100$~Hz) is limited \cite{Plattner23icccn,Yamazato16icc}.
Photodiodes offer a high temporal resolution (typically on the order of GHz), while they do not have enough spatial resolution that is necessary for separating different sources and localization \cite{Cuailean21ipc}.
Optical communication using event cameras has recently been an emerging topic in event-based vision research.
Event cameras achieve a high temporal resolution (typically up to $1$--$10 \si{\micro\second}$) and a high spatial resolution (up to $1$ Mpix), making them an ideal receiver, as shown in \cref{tab:vlccomparison}.
Also, they offer other advantages such as high dynamic range, since they output only differences of logarithmic intensities \cite{Posch08iscas, Lichtsteiner06isscc},
making them serve as ``eyes'' for robots (e.g., mobile, automotive, drones) \cite{Gallego20pami}.
\begin{figure}[t]
  \centering
  {{\includegraphics[width=\linewidth]{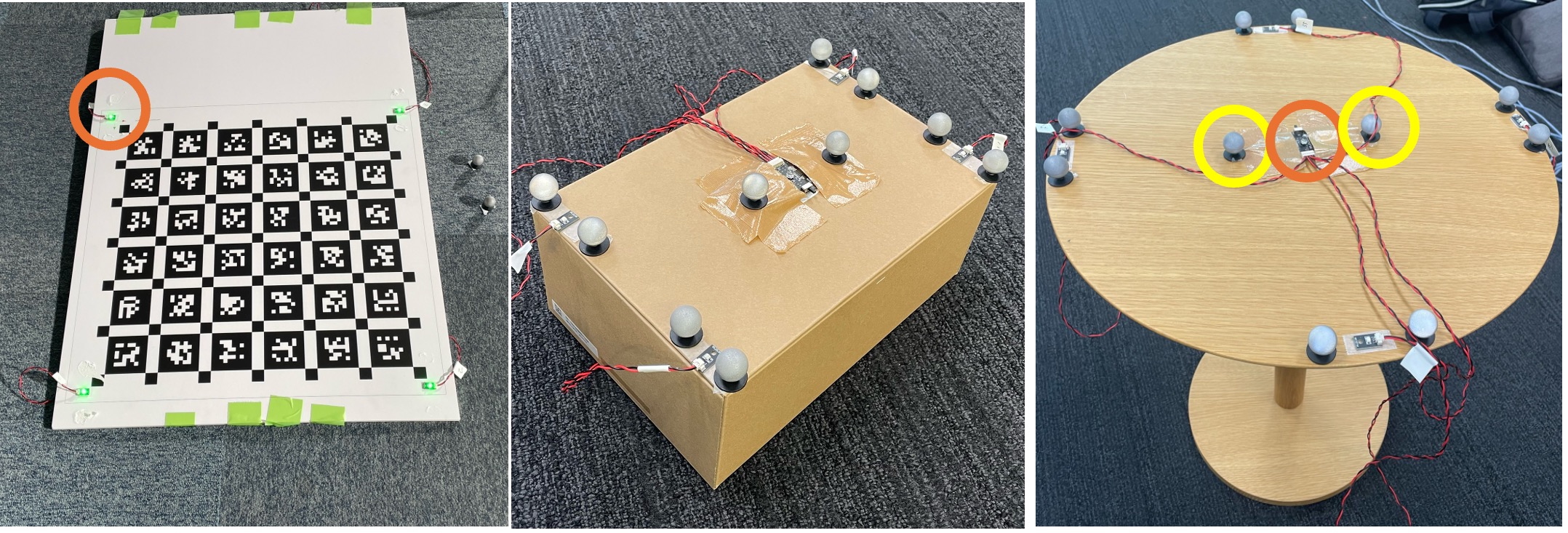}}}
\caption{LED markers on different objects. Orange circles indicate the LEDs, each of which transmits different ID signals. Yellow circles are markers for the motion capture system.}
\label{fig:methodMarker}
\end{figure}

Previous work have shown various VLC applications with event cameras on different modulation protocols.
Protocols can be categorized into four:
blink-based, frequency-based, interval-based, and change-based.
The blink-based protocol encodes the binary information as the binary ON/OFF of the illumination, i.e., blink.
Preliminary work analyzes different transmission rates (bps) at different distances in static scenes \cite{Wang22iros}.
Localization under static environments is also tested in \cite{Kitade24icra}.
The frequency-based protocol identifies LEDs with different blink frequencies.
Although it does not transmit binary information, and hence has limited communication capabilities,
it can be utilized to track distinctive LEDs with light-weight algorithms for localization of robots \cite{Salah22tim,Ebmer24wacv}.
The interval-based protocol encodes information as the interval of two consecutive ON or OFF of LEDs. 
It can be used to localize the LED markers from a stationary event camera \cite{Chen20sensors,Censi13icirs,Shen2018vnc},
as well as pan-tilt--angle estimation of a moving camera \cite{Muller11icrb}.
Recently the change-based protocol has been proposed, which encodes information by changing the intensity of LEDs (i.e., modulation) \cite{You24jrtip,Tang22iccw,Joubert19applied}.

\begin{figure*}[t]
  \centering
  {{\includegraphics[clip,trim={0 2.2cm 8.9cm 0cm},width=0.9\linewidth]{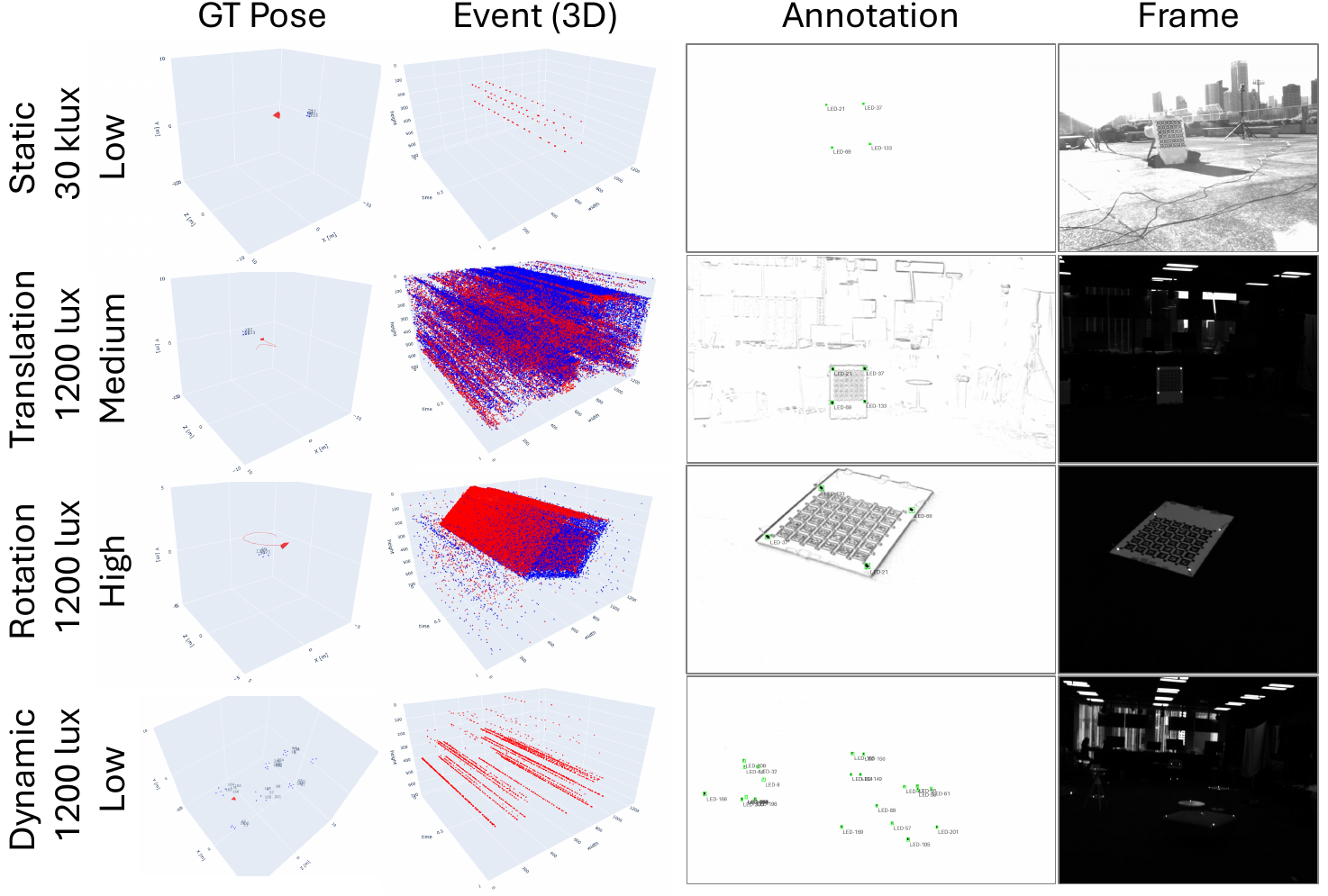}}}
\caption{\emph{Examples of the dataset.} Each row describes examples of different motion patterns, scene luminances, and camera sensitivities. The ``Low'' sensitivity sequences (first and last rows) only record markers, while others record markers and edges. Best viewed online (enlarged).}
\label{fig:methodDataset}
\end{figure*}

\section{Dataset}
\label{sec:dataset}

This section presents the details of the proposed dataset: hardware (camera and LED markers) settings (\cref{sec:dataset:hardware}),
recording scenarios (\cref{sec:dataset:sequences}),
and annotations (\cref{sec:dataset:annotation}).

\subsection{Hardware}
\label{sec:dataset:hardware}

\textbf{Receiver.}
The E-VLC dataset focuses on the scenes where the receiver (cameras) moves at a walking speed, while the scene (including LED markers) is static.
To this end, we build a custom mobile recording system equipped with two high-resolution cameras, a tablet, and a trigger box, enabling us to achieve high-quality calibration and synchronization. %
Although some cameras (e.g., DAVIS \cite{Brandli14ssc,Taverni18tcsii}) can record co-located events and frames, their data quality and pixel pitch are insufficient for precise localization purposes.
Our custom-built system consists of a high-resolution event camera (Prophesee IMX636, $1280\times720$ pix \cite{Finateu20isscc}) and a frame camera (Basler acA1300-200um, 1280$\times$1024 pix).
Both cameras have a slight baseline (i.e., stereo settings) and are recorded to a tablet (Surface 11 Pro, Ubuntu 22.02) via USB cables (see \cref{fig:methodRecording}).
The cameras are calibrated following \cite{Muglikar21cvprw}.
Notice that there are small offsets between the cameras and the markers of the motion capture, which are inevitable by design.
Event cameras are known to drop data or produce invalid timestamps when there is too much data (e.g., fast motion, flickering source), hence we choose the camera sensitivity carefully, as we report in \cref{sec:dataset:sequences}.

The trigger box sends synchronization pulses at $120$Hz to three different destinations:
the two cameras and an external motion capture (OptiTrack).
We design one of the trigger cables to be $10$~m so that it allows the recording device to move inside the motion capture system,
which provides GT positions of the camera and LED markers at sub-millimeter accuracy at $100$Hz.
The recording sites are indoor and outdoor of about $15$m $\times$ $10$m sizes.
The frames are hardware-triggered and record videos at $40$~fps.
\Cref{fig:methodRecording} shows the recording system, scenes, and example data.

\textbf{Transmitter.}
As LED markers, we use bullet-shaped LEDs (OSPG5111A, Green, $0.1$~W) with a custom microcontroller that controls the LED modulation.
One microcontroller controls five LEDs with different blink patterns, whose timings (temporal resolution) are synchronized.
The data transmitted are a few bytes of numbers (IDs), and each LED repeats the assigned blink pattern, which is typically around $10$~ms.

Several prior work have proposed modulation protocols (see \cref{sec:related}).
Among them, we choose the interval-based protocol,
since ($i$) it is robust among others tested,
($ii$) it is popularly used (e.g., supported by Prophesee's Metavision SDK \cite{MetavisionLED} as well as other prior work \cite{Chen20sensors,Censi13icirs,Shen2018vnc,Muller11icrb}),
and ($iii$) it is robust against the tail-event effect of negative events in dark scenes, since it relies on positive (ON) events only.
The time-interval protocol parameters consist of the fundamental frequency $\baseF$, the binary patterns, and the start pattern.
In this work, we set $\baseF = 5$~kHz, binary ``$0$'' as $1/\baseF$, ``$1$'' as $2/\baseF$, and start as $3/\baseF$ that are distinguished with $1/\baseF$ lighting off.

In total, we use 40 LED markers (i.e., 8 microcontrollers), all of which transmit different IDs.
The markers are attached to different objects, such as boxes and tables, as well as an Aruco marker (printed on A1 size) to benchmark with frame-based localization methods. 
\Cref{fig:methodMarker} shows examples of LED markers and objects.
The precise LED positions are measured with the motion capture system before the camera recordings.

\subsection{Scenarios}
\label{sec:dataset:sequences}

The details of the sequences, conditions, and data statistics are summarized in \cref{tab:dataset}.
In total, the dataset consists of $110$ sequences that include $7.5$G events, $88$k frames, and $266$k annotations.

\textbf{Motion type.}
We prepare four different camera motions:
\emph{static}, \emph{translation}, \emph{rotation}, and \emph{dynamic}.
All motions except for \emph{static} are at a normal walking speed with the hand-held recorder.
\Cref{fig:methodDataset} shows examples of the data.

\emph{Static} sequences use four markers that are attached to the corner of the AprilGrid (see \cref{fig:methodMarker} (left) for the markers).
During the recording, the device is mounted on a tripod and does not move.
We test different locations with different distances and angles from the marker against a wall, spanning from $2$~\si{\meter} to $14$~\si{\meter}.
\emph{Translation} sequences use the same four LEDs and the Aruco marker,
however, the recording device moves with mainly translational motion.
Some sequences focus on the translation along the camera's optic axis (i.e., perpendicular to the markers), to cover practical cases for mobile applications,
while others include the translation along the other directions (i.e., parallel to the markers).
In \emph{rotation} sequences, the same markers are located on the floor, and the camera motion is mainly rotation, focusing on the marker in the field of view.
Contrary to the other three patterns that use four LEDs, \emph{dynamic} sequences use $35$ LED markers that are attached to different objects, such as boxes and tables as shown in \cref{fig:methodMarker}.
The recording device moves freely among the objects and receives signals from partially visible LEDs.

\begin{table}[t!]
\centering
\caption{Event camera sensitivity settings.}
\setlength{\tabcolsep}{4pt}
\begin{tabular}{l*{4}{S}}
\toprule
 & \text{Bias (OFF)} 
 & \text{Bias (ON)} 
 & \text{High-pass filter} 
 \\ 
\midrule
\text{Low} & \text{180} & \text{60} & \text{120} \\
\text{Medium} & \text{70} & \text{60} & \text{60} \\
\text{High} & \text{45} & \text{45} & \text{30} \\
\bottomrule
\end{tabular}
\label{tab:dataset:sensitivity}
\vspace{-2ex}
\end{table}

\begin{figure*}[t]
  \centering
  {{\includegraphics[clip,trim={0 7cm 8cm 0},width=0.9\linewidth]{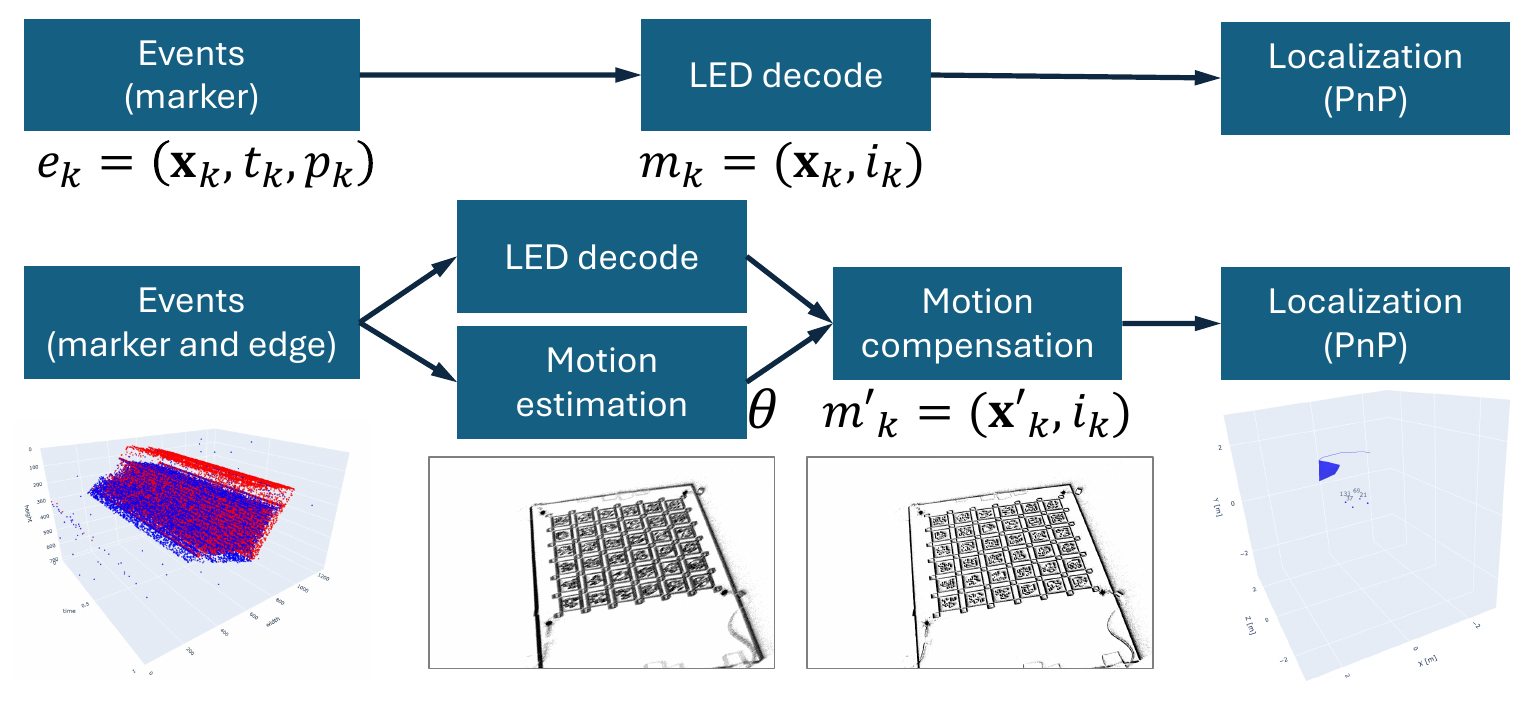}}}
\caption{\emph{Proposed motion compensation pipeline}. With the low camera sensitivity (the top diagram), LED information is decoded directly from events. With the middle or high sensitivities (bottom diagram), we leverage the motion estimation and compensation algorithms to improve the localization accuracy.}
\label{fig:methodCompensation}
\end{figure*}

\textbf{Camera sensitivity.}
Since the LED markers emit high-frequency intensity changes,
one could configure the event camera parameters (e.g., biases, high-pass filters) so that it only records such signals, discarding low-frequency events generated by moving edges in the scene.
However, in practical use-cases of event cameras for VLC, it is important for the camera to receive both events from such moving edges and LEDs, so that the event data can be utilized in other computer vision tasks, such as motion estimation, object detection, and deblur. %
However, since LEDs generate lots of events and the cameras cannot guarantee proper timestamps beyond a certain amount of data (typically tens to hundreds of Mega-events per second), we carefully select three different parameters to record valid data.
\Cref{tab:dataset:sensitivity} describes the parameters tested:
the \emph{low} sensitivity is tuned best for the LED signals and rarely receives other events, the \emph{high} sensitivity captures both LEDs and scene edges, and the \emph{medium} sensitivity is in the middle of the two.

\textbf{Scene luminance.}
We also provide sequences in various scene luminances to leverage the high-dynamic range of event cameras.
The data is recorded both outside (on a sunny day) and inside, where we can control the luminance.
The luminance conditions are among $10$, $600$, $1200$, and $30$k (bright outside), depending on sequences (see \cref{tab:dataset}).

\subsection{Annotations}
\label{sec:dataset:annotation}
After recording, we annotate bounding boxes of LED pixels with the IDs for the event camera data as shown in \cref{fig:methodDataset} (the third column).
The annotation frames are based on the trigger signals (i.e., $120$Hz).
The LED light may cause glare in the event camera, especially when the camera is close enough to the light source, in which case the observed events form sharp tails from the center of the LED.
We include such tails in the bounding boxes.

\section{Methodology}
\label{sec:method}

In this section, we briefly summarize the LED detection and pose estimation from events in \cref{sec:method:detection} and propose the motion compensation pipeline in \cref{sec:method:motionCompensation}.

\subsection{LED detection and pose estimation}
\label{sec:method:detection}

An event camera captures intensity changes at each pixel and generates a stream of events $e_k \doteq (\bx_k, t_k, \pol_k)$, where $(\bx_k, t_k)$ is space-time coordinates and polarity $\pol_k \in \{+1, -1\}$ is the sign of the change.
An event is generated when the logarithmic intensity $L$ at a pixel $\bx$ changes by a certain threshold value $C$, i.e., 
$\Delta \Lum(\bx_k,t_k) \doteq \Lum(\bx_k,t_k) - \Lum(\bx_k, t_k-\Delta t_k) = \pol_k \, C$.

Although the intensity change due to the marker may be much larger than $C$,
and hence an LED change (e.g., OFF to ON) may generate more than one event, we simply ignore such cases and focus on the time interval of events that are close enough to the base frequency of the LED blink.
To this end, we use the time-map representation \cite{Delbruck08issle,Lagorce17pami} that stores the timestamp of a previous event at the same pixel $\bx$.
The time map $\cT(\bx_k) = t_k$ is updated using the timestamp of the new event $t_{k+1}$, if the time interval is larger than the previous time-map at the pixel $\bx_{t+1}$ with a tolerance $\tau$,
\begin{equation}
\label{eq:timeMapUpdate}
\cT (\bx_{k+1}) = t_{k + 1}, \text{if } t_{k + 1} > \cT (\bx_{k+1}) + \frac{1}{\baseF} - \tau.
\end{equation} 

The time-map provides the history of the time intervals of previous events, and we decode the marker IDs based on the series of the previous time-map since the start signal.
After detecting and decoding LEDs, we use SQPNP \cite{Terzakis20eccv} to estimate the camera pose, using the calibration parameters. %
Since our setting relies on the small number of keypoints (typically less than 10 pixels),
we find that RANSAC does not contribute to stabilizing the pose estimation and hence we do not use it.

\subsection{Motion Estimation and Compensation}
\label{sec:method:motionCompensation}

Since the event camera moves, we observe that the detection accuracy degrades along with the displacement of the LED pixels in the image plane.
One might use tracker algorithms, however, this requires integration with the time map representation, which we leave as future work.
In this work, we propose to utilize motion estimation and compensation to improve the decoding and localization accuracy.

Contrast Maximization (CMax) \cite{Gallego18cvpr} is a state-of-the-art method to estimate various motion from events alone,
such as rotation \cite{Gallego17ral,Kim21ral,Shiba22aisy,Gu21iccv,Guo24tro},
feature flow \cite{Zhu17icra,Zhu17cvpr,Shiba22sensors,Seok20wacv,Stoffregen19cvpr},
and dense optical flow \cite{Shiba22eccv,Shiba24pami}.
In CMax, events $\cE \doteq \{e_k\}_{k=1}^{\numEvents}$
are assumed being caused by moving edges (i.e., brightness constancy assumption),
and are geometrically warped %
to $\cE'_{\tref} \doteq \{e'_k\}_{k=1}^{\numEvents}$ at a reference time $\tref$ using a motion model $\Warp$:
\begin{equation}
\vspace{-0.5ex}
e_k \doteq (\bx_k,t_k,\pol_k) \;\,\mapsto\;\, 
e'_k \doteq (\bx'_k,\tref,\pol_k).
\vspace{-0.5ex}
\end{equation}
Then, the warped events are aggregated on an image of warped events (IWE):
\begin{equation}
\label{eq:IWE}
\vspace{-0.5ex}
\textstyle
\IWE(\bx; \cE'_{\tref}, \bparams) \doteq \sum_{k=1}^{\numEvents} \delta (\bx - \bx'_k),
\vspace{-0.5ex}
\end{equation}
as the histogram of the warped events $\bx'_k$.  %
The Dirac delta is approximated by a Gaussian, 
$\delta(\bx-\bmu)\approx\cN(\bx;\bmu,\epsilon^2\mId)$ with $\epsilon=1$ pixel.
Next, an objective function, such as the image contrast of the IWE~\eqref{eq:IWE} (e.g., \cite{Gallego19cvpr})
measures the goodness of fit between the events and the warp parameter.
As the warp models in this work, we use two types:
feature flow $\bx' = \bx + (t -t_\text{ref}) \bparams$ with a constant image velocity $\bparams \equiv (v_x, v_y)^{\top}$ for translational motion,
and 
$\bx^{h\prime} \sim \Rot(t \angvel)\,\bx^h$
with an angular velocity $\bparams \equiv (\omega_{x},\omega_{y},\omega_{z})^{\top}$,
calibrated homogeneous coordinates $\bx^h$,
and exponential coordinates $\Rot(\bphi) \doteq \exp(\bphi^\wedge)$ \cite{Barfoot15book,Gallego14jmiv} for rotational motion.

Our proposed pipeline is summarized in \cref{fig:methodCompensation}.
Thanks to the parameter settings, we utilize the event information from scene edges (non-LEDs) to estimate the camera motion.
After the motion estimation, we warp the events and align its spatial coordinates, which we feed to the downstream LED detection and localization pipeline (\cref{sec:method:detection}).
Furthermore, despite the brightness constancy assumption of the CMax framework, we find that CMax works well to estimate motion on low-sensitivity event data (i.e., events mainly from LED markers), which breaks the assumption.

\section{Experiments}
\label{sec:experim}

\textbf{Metrics and baseline.}
For future benchmarking purposes, we provide the baseline results of the marker detection and localization.
The metrics used to assess the LED decoding are detection rate (DR),
which is the ratio of bounding boxes that have at least one pixel with the correct marker ID decoding.
For the pose metrics, we report the $L_1$ error of the location.
We compare the results between the proposed motion compensation (``Ours'') and without it (``Baseline'').

\textbf{Hyper-parameters.}
The motion estimation is based on the Newton-CG optimization algorithm with a maximum of $50$ iterations.
The evaluation is at the frame timings (i.e., $40$fps).
We use the 3-DOF rotational motion for the rotation sequences and the 2-DOF feature flow for others.

\begin{figure}[t]
  \centering
  {{\includegraphics[clip,trim={0 7.3cm 5cm 0},width=\linewidth]{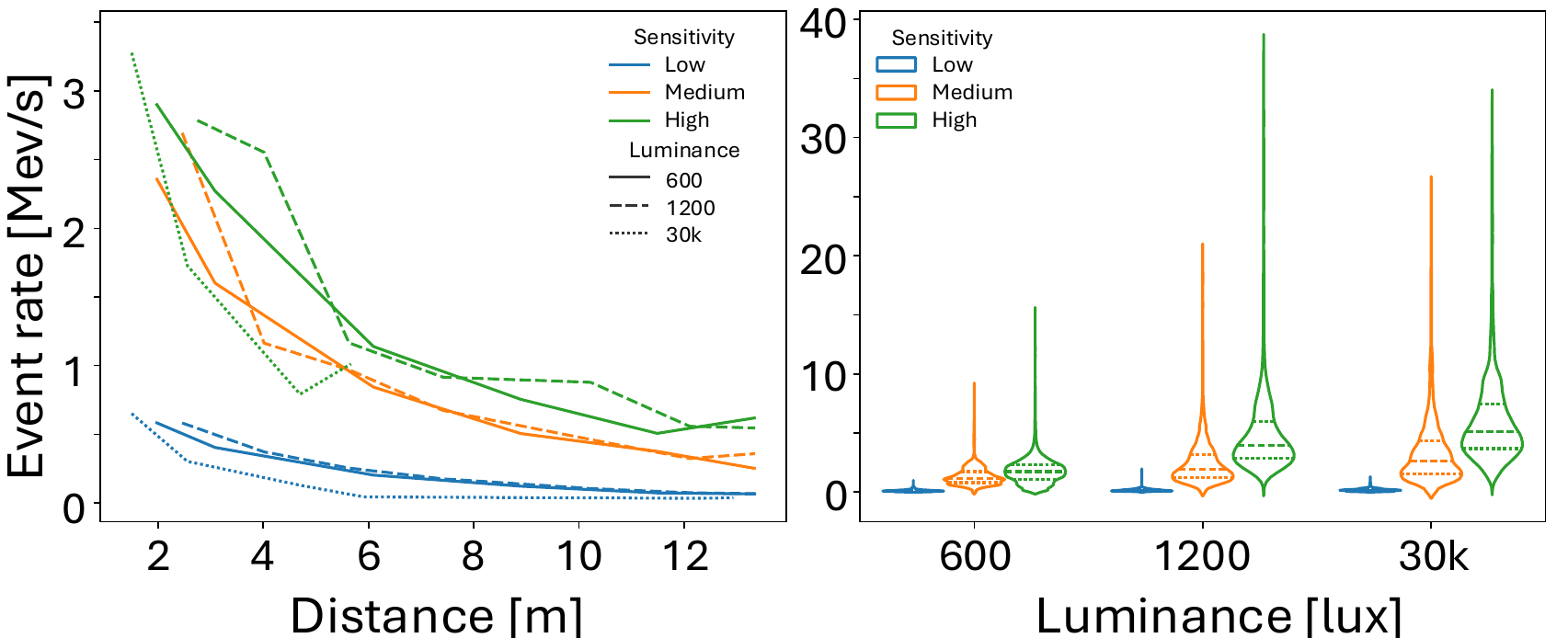}}}
\caption{Event rate analysis. (left) The \emph{static} sequences mainly consist of events from the LED markers. (right) The \emph{dynamic} sequences consist of events from the LED markers and the camera motion, depending on the camera sensitivity.
}
\label{fig:analysis:eventRate}
\end{figure}

\textbf{Event-rate analysis.}
Event cameras need careful bias configurations when the scene includes flickering light sources due to possibly large amounts of data, as mentioned in \cref{sec:dataset:sequences}.
Here, we first provide further detailed analysis both for the simplest sequences (i.e., \emph{static}) and the complex sequences (i.e., \emph{dynamic}).
\Cref{fig:analysis:eventRate} shows the event rate in the unit of Mev/s (Mega-event per second) for the static and dynamic sequences.
The static sequences do not expect events from the camera motion, regardless of the sensitivity.
The camera responds to the four LED markers, whose sizes in the image plane increase at closer distances.
The dynamic sequences generate events from the markers and the camera motion, depending on the sensitivity.
In all scene luminance conditions,
the \emph{high} sensitivity results in the largest amount of events,
and we observe the camera drops data for even higher sensitivity (e.g., lower contrast threshold) or faster camera motion.

\subsection{Comparison Between Events And Frames}
\label{sec:experim:frameVsEvent}

\begin{figure}[t]
  \centering
  {{\includegraphics[clip,trim={0 4.6cm 19.5cm 0},width=\linewidth]{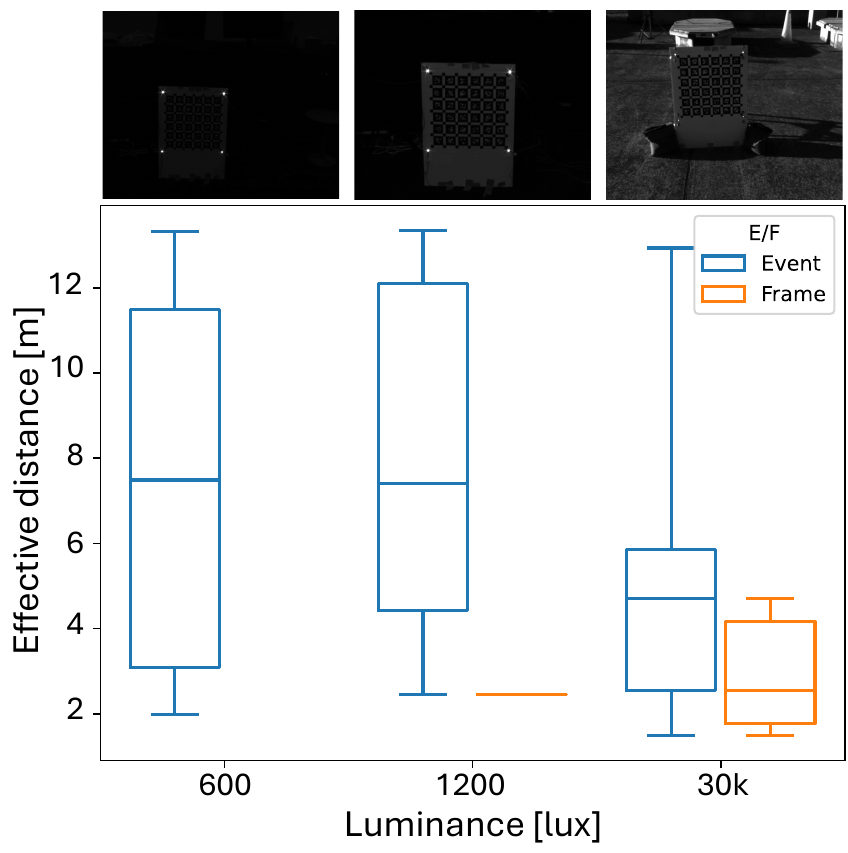}}}
\caption{\emph{Effective marker detection distances for frames and events}. Frame data collapse as the scene becomes darker, limiting the detection range, while event data do not suffer. 
}
\label{fig:analysis:effectiveDistance}
\vspace{-2ex}
\end{figure}

\textbf{Effective distance of localization.}
\Cref{fig:analysis:effectiveDistance} shows the effective distances of camera localization, comparing the LED-based (event-based) and the Aruco-based (frame-based) methods.
The frame-based method can not detect the marker or localize in dark scenes (e.g., $600$ lux).
The effective distance for frames becomes shorter under $1200$ lux than under $30$k lux,
however, the effective distance for events does not change significantly due to the scene brightness,
clearly showcasing the advantage of LED-based localization.

\begin{table}[t]
\centering
\caption{\emph{Comparison between frame-based localization and event-based localization} using rotation sequences.}
\label{tab:arucoVsEvent}
\adjustbox{max width=\linewidth}{%
\setlength{\tabcolsep}{3pt}
\begin{tabular}{l*{3}{S[table-format=1.3,round-mode=places,round-precision=3]}}
\toprule
& \text{Detected frames $\uparrow$} & \text{Mean [m] $\downarrow$} & \text{Median [m] $\downarrow$}
\\
\midrule
Event (H) & \textbf{4185} & 0.1837827717 & 0.1808599722 \\
Frame & \text{1038} & \bnum{0.1708550848} & \bnum{0.1775895286} \\
 \midrule
Event (M) & \textbf{3236} & 0.1958224683 & 0.1908684359 \\
Frame & \text{755} & \bnum{0.1724696516} & \bnum{0.1767339713} \\
 \midrule
Event (L) & \textbf{2871} & 0.2038843989 & 0.197312718 \\
Frame & \text{374} & \bnum{0.1706582271} & \bnum{0.1730221357} \\
\bottomrule
\end{tabular}
}
\end{table}

\begin{table*}[t]
\centering
\caption{Results of LED detection and localization.}
\label{tab:detection}
\adjustbox{max width=\linewidth}{%
\setlength{\tabcolsep}{3pt}
\begin{tabular}{ll*{13}{S[table-format=1.3,round-mode=places,round-precision=3]}}
\toprule
\multicolumn{2}{c}{Sensitivity}
 & \multicolumn{4}{c}{High} 
 & \multicolumn{4}{c}{Medium}
 &\multicolumn{4}{c}{Low} \\
 \cmidrule(l{1mm}r{1mm}){3-6}
 \cmidrule(l{1mm}r{1mm}){7-10}
 \cmidrule(l{1mm}r{1mm}){11-12}
 \cmidrule(l{1mm}r{1mm}){13-14}
\multicolumn{2}{c}{Luminance}
 & \multicolumn{2}{c}{1200} 
 & \multicolumn{2}{c}{600} 
 & \multicolumn{2}{c}{1200} 
 & \multicolumn{2}{c}{600} 
 & \multicolumn{2}{c}{1200} 
 & \multicolumn{2}{c}{600} \\ 
 \cmidrule(l{1mm}r{1mm}){3-4}
 \cmidrule(l{1mm}r{1mm}){5-6}
 \cmidrule(l{1mm}r{1mm}){7-8}
 \cmidrule(l{1mm}r{1mm}){9-10}
 \cmidrule(l{1mm}r{1mm}){11-12}
 \cmidrule(l{1mm}r{1mm}){13-14} 
 & & \text{DR $\uparrow$} & \text{Pose [m] $\downarrow$}
 & \text{DR $\uparrow$} & \text{Pose [m] $\downarrow$}
 & \text{DR $\uparrow$} & \text{Pose [m] $\downarrow$}
 & \text{DR $\uparrow$} & \text{Pose [m] $\downarrow$}
 & \text{DR $\uparrow$} & \text{Pose [m] $\downarrow$}
 & \text{DR $\uparrow$} & \text{Pose [m] $\downarrow$}
\\
\midrule
\multirow{2}{*}{Translation(Z)} & Baseline
& 0.9077066501 & 1.10492668 & 0.9661843876 & 1.093214854
& 0.9034400545 & 1.168473458 & 0.9645856663 & 1.233726022
& 0.9774293591 & 0.9342335744 & 0.9908060453 & 1.035975547
\\
& Ours
& \bnum{0.9125233064} & \bnum{0.784852701} & \bnum{0.9685397039} & \bnum{0.7821393613}
& \bnum{0.9148501362} & \bnum{0.7526780461} & \bnum{0.9697648376} & \bnum{0.8582434511}
& 0.9774293591 & \bnum{0.6990411547} & 0.9909319899 & \bnum{0.8071990698}
\\
\midrule
\multirow{2}{*}{Translation(XYZ)} & Baseline
& \novalue & \novalue & \novalue & \novalue 
& 0.9777424483 & 0.8342718995 & 0.9499103541 & 0.7953522011
& 0.9669332813 & 0.8523976825 & 0.9686477644 & 0.6731176681 
\\
& Ours
& \novalue & \novalue & \novalue & \novalue 
& \bnum{0.9806240064} & \bnum{0.7363496693} & \bnum{0.9524876737} & \bnum{0.6586212096}
& \bnum{0.9671283652} & \bnum{0.6586212096} & \bnum{0.9701472192} & \bnum{0.5623945983} 
\\
\midrule
\multirow{2}{*}{Rotation} & Baseline 
& 0.9241322702 & 0.2868101184 & 0.9831349206 & 0.2395641917
& 0.9728812452 & 0.2188860868 & 0.9775205806 &0.2542872744
& 0.9883333881 & 0.2273962521 & 0.9874683031 &0.250309141
\\
& Ours
& \bnum{0.9290572233} & \bnum{0.2470065617} & \bnum{0.9831349206} & \bnum{0.1886081945}
& \bnum{0.9749251766} & \bnum{0.1767903052} & \bnum{0.980474517} & \bnum{0.206061486}
& \bnum{0.9887623204&} \bnum{0.1746239852} & \bnum{0.9876320228} & \bnum{0.2108162178}
\\
\bottomrule
\end{tabular}
}
\vspace{-2ex}
\end{table*}

\textbf{Localization accuracy.}
\Cref{tab:arucoVsEvent} reports the comparison between frame-based and event-based localization on the \emph{rotation} sequences ($1200$ lux), where the marker size is $0.5$~m and the radius of the motion is roughly $2$--$3$m.
The accuracies between events (LED markers) and frames (Aruco markers) are similar, confirming the effectiveness of the LED-based localization.
We observe slight ($1$--$3$~cm) performance improvement with frames, which can be attributed by the number of markers used ($36$ markers, compared with $4$ LEDs).
Notice that LED detection is by far more successful than Aruco marker detection (i.e., ``Detected frames''),
which is due to motion blur in the frames.
In conclusion, both results in \cref{fig:analysis:effectiveDistance,tab:arucoVsEvent} evidence that the LED-based localization works at larger distance and under various scene illumination,
while maintaining the accuracy that is expected in the conventional frame-based method.

\subsection{Results of LED Detection and Localization}
\label{sec:experim:detectAndLocalize}

\begin{figure}[t]
  \centering
  {{\includegraphics[clip,trim={0 9.1cm 18.8cm 0},width=\linewidth]{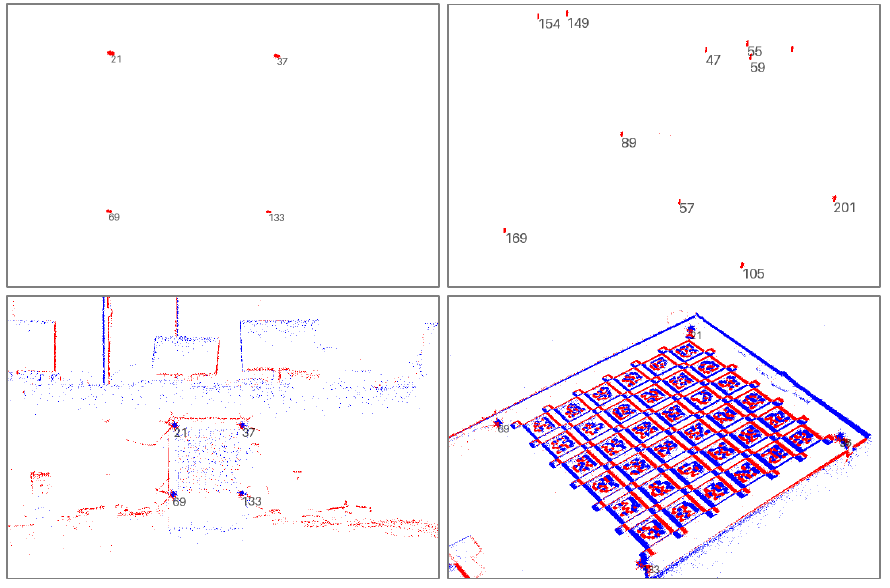}}}
\caption{\emph{Detection results} for different sequences. (Top) Low sensitivity, (Bottom) medium and high sensitivity.}
\label{fig:result:detection}
\end{figure}

\begin{figure}[t]
  \centering
  {{\includegraphics[clip,trim={0 0.4cm 19.2cm 0},width=\linewidth]{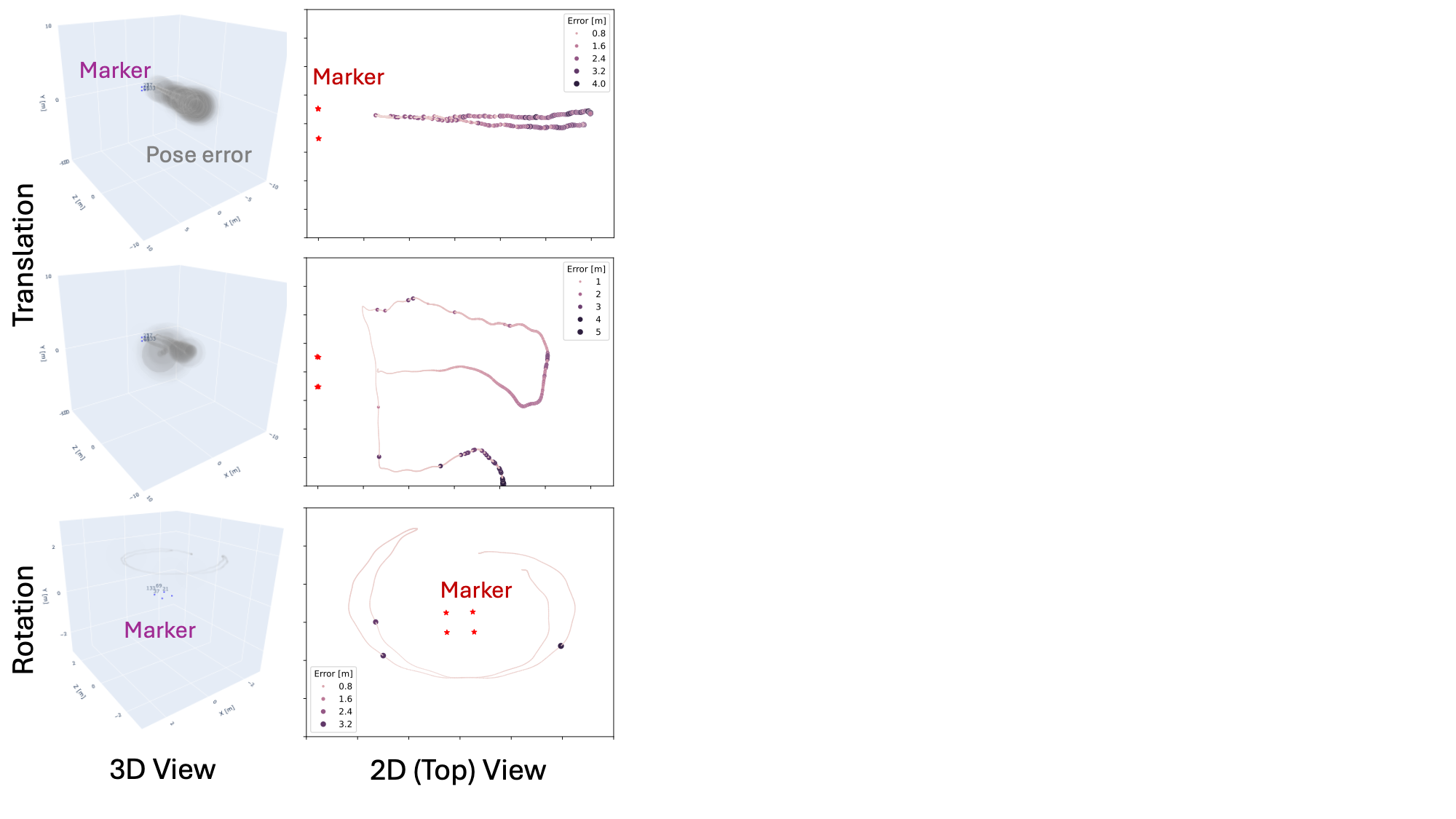}}}
\caption{Results of localization errors.}
\label{fig:result:localization}
\vspace{-3ex}
\end{figure}

The qualitative results for various sequences are shown in \cref{fig:result:detection} for detection and in \cref{fig:result:localization} for localization, respectively.
We confirm that LEDs are successfully detected and decoded in most of the sequences.
\Cref{tab:detection} reports the quantitative results of the detection rate, showing it achieves over $90$\% for all sequences.
The detection fails under large pixel displacements, i.e., large motion of the camera.

The localization accuracy is also reported in \cref{tab:detection}.
Our proposed motion compensation pipeline constantly achieves smaller errors than the baseline.
Notice that the translation sequences consist of various distances ($2$--$15$~m) between the camera and the markers,
which explains the relatively larger error (about $80$~cm) compared with the rotation sequences (about $20$~cm).
The qualitative results (\cref{fig:result:localization}) show examples of the different errors based on the distance from the marker.
We observe no significant differences in the pose errors among different camera sensitivities or scene illuminations.

\section{Limitations}
\label{sec:limitations}

While we find that other light sources (i.e., other flickering light sources than LED markers) are not a problem due to the marker's specific blink patterns,
reflective surfaces in the scene may interfere the detection and localization.
In practical applications, a more robust algorithm for localization is desirable, e.g., utilizing RANSAC.

The blink protocol does not have error corrections such as parity bits, hence it is important to explore a more robust protocol.
We observe the latency of negative events increases under low illuminations (i.e., ``tail'' effect).
This is reported to be specific to the sensor used (IMX636),
hence different cameras could improve the error rate in the future.

In this work, we use simple motion models, such as feature flow and rotational motion, which the recording scenarios do not always fit.
It would be desirable to combine SLAM (simultaneous localization and mapping) with the LED marker settings.
Also, the proposed motion estimation and compensation increase the computation, hence it is important to benchmark the runtime.

\section{Conclusion}
\label{sec:conclusion}

This work presents the first public dataset (E-VLC) using modulated LEDs for event cameras, which consists of precisely synchronized events, frames, ground-truth pose of LED markers and cameras, and bounding-box annotations for LEDs in various recording conditions (motion type, scene luminance, event camera sensitivity) to cover real-world applications.
We also propose a novel method using motion compensation based on Contrast Maximization that improves localization.
We hope that the proposed dataset serves as a future benchmark for both motion-related classical computer vision tasks and LED marker decoding tasks simultaneously,
broadening event camera applications.

{
    \small
    \bibliographystyle{ieeenat_fullname}
    \bibliography{all,references}
}

\end{document}